\documentclass[conference]{IEEEtran}
\IEEEoverridecommandlockouts
\usepackage{spconf}
\usepackage{cite}
\usepackage{amsmath,amssymb,amsfonts}
\usepackage{graphicx}
\usepackage{textcomp}
\usepackage{xcolor}
\usepackage{url}
\usepackage{multirow}
\usepackage{graphicx}
\def\BibTeX{{\rm B\kern-.05em{\sc i\kern-.025em b}\kern-.08em
    T\kern-.1667em\lower.7ex\hbox{E}\kern-.125emX}}
\usepackage{placeins}
\usepackage{float}
\usepackage{hyperref}
\usepackage{tabularx,colortbl}
\usepackage{ifthen}
\usepackage{caption,subcaption}
\usepackage{booktabs}
\renewcommand{\thetable}{\Roman{table}}

\usepackage{enumitem}
\usepackage{stfloats}
\usepackage{wrapfig}
\usepackage{verbatim}
\usepackage{tabularx}
\usepackage{booktabs}
\usepackage{multirow}
\usepackage{caption}
\usepackage{flushend}
\usepackage{algpseudocode}
\usepackage{cleveref} 
\usepackage{titlesec}
\usepackage{pifont,booktabs,adjustbox}
\titlespacing*{\subsection}{0pt}{3pt}{1pt}
\usepackage[T1]{fontenc}
\usepackage{mathptmx} 
\usepackage{epstopdf} 

\begin{document}

\title{\textit{RMT-KD}: Random Matrix Theoretic Causal Knowledge Distillation}

\def\blind{0} 

\name{Davide Ettori$^1$, Nastaran Darabi$^1$, Sureshkumar Senthilkumar$^1$, and Amit Ranjan Trivedi$^1$ \vspace{-10pt}} 
\address{$^1$University of Illinois Chicago, Chicago, IL\\
\texttt{detto3@uic.edu, amitrt@uic.edu}}

\maketitle

\begin{abstract}
Large deep learning models such as BERT and ResNet achieve state-of-the-art performance but are costly to deploy at the edge due to their size and compute demands. We present \textit{RMT-KD}, a compression method that leverages Random Matrix Theory (RMT) for knowledge distillation to iteratively reduce network size. Instead of pruning or heuristic rank selection, RMT-KD preserves only informative directions identified via the spectral properties of hidden representations. RMT-based causal reduction is applied layer by layer with self-distillation to maintain stability and accuracy. On GLUE and CIFAR-10, RMT-KD achieves up to 80\% parameter reduction with only 2\% accuracy loss, delivering 2.8$\times$ faster inference and nearly halved power consumption. These results establish RMT-KD as a mathematically grounded approach to network distillation.
\end{abstract}

\vspace{5pt}

\section{Introduction and Prior Works}
Recent advances in natural language processing (NLP) and computer vision rely on increasingly large deep learning models such as BERT \cite{devlin2019bert} and convolutional networks such as ResNet \cite{he2016resnet}. While highly accurate, these models incur substantial inference latency, memory footprint, and carbon emissions \cite{strubell2019energy}. Classical compression methods include knowledge distillation (KD), which transfers predictions from a frozen teacher to a smaller student \cite{hinton2015distillation}, with refinements such as DistilBERT \cite{sanh2019distilbert} and progressive shrinking \cite{cai2020onceforall}; pruning, which removes unimportant weights or channels \cite{han2016deepcompression}; and low-rank factorisation, which compresses convolutional filters \cite{jaderberg2014lowrank} and adaptation matrices in large language models (LLMs) \cite{hu2022lora}. \textit{Yet} these techniques often depend on heuristic thresholds, yield hardware-unfriendly sparsity, or lack a principled statistical rule.

In this paper, we ask: \textit{Is there a principled statistical basis for model reduction?} We argue that Random Matrix Theory (RMT) provides such a foundation \cite{bun2017rmtreview}. In high-dimensional settings, the eigenvalue spectrum of activation covariances typically separates into a bulk that reflects random noise and a small number of outliers that capture structured, causal features \cite{papyan2020powerlaw,martin2021implicit}. This separation, formalized by the Marchenko–Pastur (MP) law \cite{marchenko1967, ettori2025eigentrackspectralactivationfeature}, offers a rigorous criterion for identifying informative directions.

Building on this insight, we propose an iterative distillation framework: once the network has learned meaningful features, we extract a calibration subset, estimate activation covariances, and project onto the causal subspace defined by outlier eigenvalues. The resulting models are narrower yet fully dense and deployable on standard hardware without sparse operations. After each reduction, self-distillation from the previous checkpoint restores accuracy and prevents catastrophic forgetting. Repeating this cycle across layers enables progressive, RMT-guided compression.

\begin{figure}[t!]
    \centering
    \includegraphics[width=\linewidth]{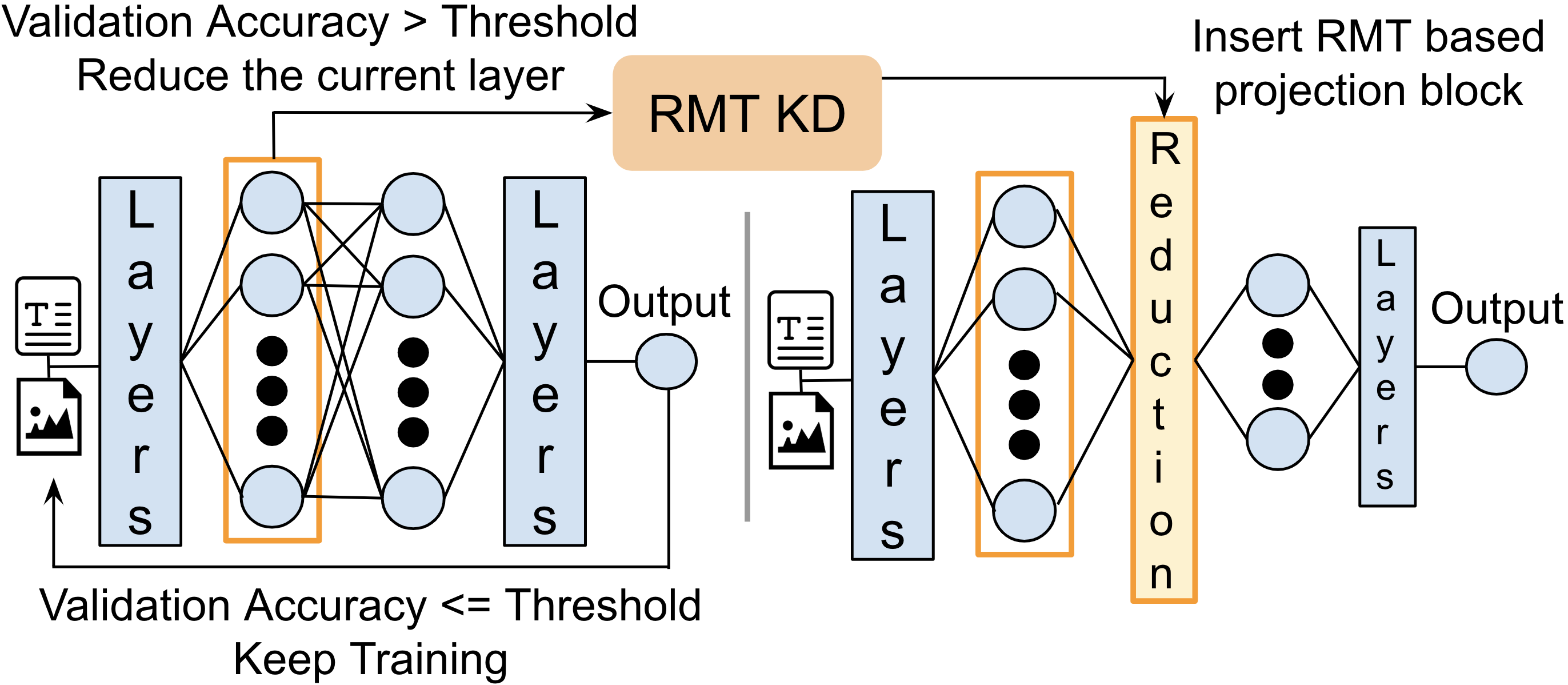}
    \caption{Architecture of RMT-KD for iterative distillation. At each stage, hidden layer activations are analyzed with RMT principles to identify causal directions, followed by projection and self-distillation. This process is repeated across layers until the benefits saturate.}\vspace{-10pt}
    \label{fig:architecture}
\end{figure}

This RMT-guided distillation is scalable across architectures and modalities, adapting projections to the spectral structure of each layer. Compression can be terminated based on validation accuracy, retained outliers, or target reductions. Our contributions are: (i) a principled, causal rule for layer reduction via RMT-based eigenvalue filtering, thereby avoiding heuristic rank selection; (ii) a model-agnostic algorithm alternating RMT projection with self-distillation; and (iii) an empirical study demonstrating state-of-the-art accuracy–efficiency trade-offs on GLUE and CIFAR-10, with significant parameter and energy savings.

\section{RMT-KD: Methodology}

We propose an iterative self-distillation method that compresses neural networks by progressively reducing their width under RMT guidance. Unlike standard knowledge distillation, where a student learns from a fixed teacher, our approach uses a single model \textit{that distills itself} at intermediate checkpoints. Each iteration treats the current model as the teacher and its reduced counterpart as the student, trained with a combined loss:
\[
L = \alpha \,\mathrm{CE}_{\text{task}} + (1-\alpha)\,\mathrm{KL}(p_{\text{old}} \,\|\, p_{\text{new}}),
\]
where $\mathrm{CE}_{\text{task}}$ is the cross-entropy loss and $\mathrm{KL}(p_{\text{old}} \,\|\, p_{\text{new}})=\sum_{i} p_{\text{old}}(i) \log \tfrac{p_{\text{old}}(i)}{p_{\text{new}}(i)}$ is the distillation loss. This regularizes training, enforcing similarity between consecutive models and mitigating catastrophic forgetting.

\vspace{5pt}
\noindent\textbf{RMT-Guided Projection:}  
Training proceeds until validation accuracy exceeds a threshold, after which a calibration subset $\mathcal{D}_{\text{cal}}$ (10\% of the training set) is used to extract hidden activations $X \in \mathbb{R}^{d \times n}$ from a target layer. This subset provides a stable snapshot of the model’s learned features without requiring the full dataset. The empirical covariance $\Sigma = \tfrac{1}{n} XX^\top$ yields spectrum $\{\lambda_i\}_{i=1}^d$, which under the spiked covariance model \cite{johnstone2001distribution, paul2007asymptotics} follows the MP law \cite{marchenko1967}:
\[
\rho_{\text{MP}}(\lambda) = \frac{1}{2\pi \lambda q \sigma^2} \sqrt{(\lambda_+ - \lambda)(\lambda - \lambda_-)}, \quad \lambda \in [\lambda_-,\lambda_+],
\]
with bulk edges $\lambda_{\pm} = \sigma^2(1 \pm \sqrt{d/n})^2$, where $q=d/n$. The noise variance $\sigma^2$ is initialized as the median eigenvalue and refined by minimizing the $\ell_2$ distance between the empirical histogram and the MP distribution. Adjusting the initialization quantile controls compression aggressiveness: higher quantiles increase $\lambda_+$ and prune more directions.

Eigenvalues $\lambda_i > \lambda_+$ identify informative causal directions \cite{ darabi2025eigenshieldcausalsubspacefiltering}. Their eigenvectors form a projection $P \in \mathbb{R}^{k \times d}$, defining a lower-dimensional causal subspace. A fixed linear layer applies $P$, and downstream layers are resized to $k < d$. Training then resumes with loss $L$, defined above. This cycle repeats across layers until compression targets are met or validation accuracy falls below the threshold. In BERT, projections are applied to embeddings; in ResNet, to convolutional channels. Fig.~\ref{fig:architecture} illustrates the workflow: \emph{train $\rightarrow$ analyse $\rightarrow$ project $\rightarrow$ fine-tune}, thereby producing compact, dense models without sparse operations which are efficient for GPU acceleration.

\vspace{5pt}
\noindent\textbf{Complexity and Scalability Analysis:}  
RMT-KD adds only lightweight spectral analysis. For a layer of width $d$ and calibration set size $n$, forming the covariance $\Sigma \in \mathbb{R}^{d \times d}$ costs $\mathcal{O}(nd^2)$ and eigen-decomposition $\mathcal{O}(d^3)$. Since $n$ is a small fraction (10\%) of training data and $d$ is bounded by hidden width, this cost is negligible relative to training. The projection adds a fixed linear layer of size $k \times d$, similar to a pruning mask but without sparsity.  

Overall, the loop scales as $\mathcal{O}(Ld^3)$ for $L$ layers, dominated by decompositions that parallelize well on GPUs. Unlike iterative pruning, which requires repeated masks and sparse kernels, RMT-KD preserves dense tensors and hardware efficiency. The cubic cost remains tractable as only a few layers are reduced per iteration, with gains compounding across layers. The method thus applies to both backbones (ResNet-50) and transformers (BERT-base), and can extend to billion-parameter LLMs via block-wise decomposition or randomized eigensolvers.

\vspace{5pt}
\noindent\textbf{Experimental Setup:}  
We trained all models from scratch: a 12-layer Transformer identical to BERT-base (139M parameters), its 6-layer TinyBERT counterpart (44M), and ResNet-50 (23M). Experiments ran on a single NVIDIA RTX~6000 GPU with CUDA, with GPU power measured using vendor tools. Datasets followed standard splits, preprocessing and tokenization: GLUE for language tasks, and CIFAR-10 for vision. For BERT, reported results on GLUE refer to the average performance across SST-2, QQP, and QNLI datasets. 

\begin{figure}[t!]
    \centering
    \includegraphics[width=\linewidth]{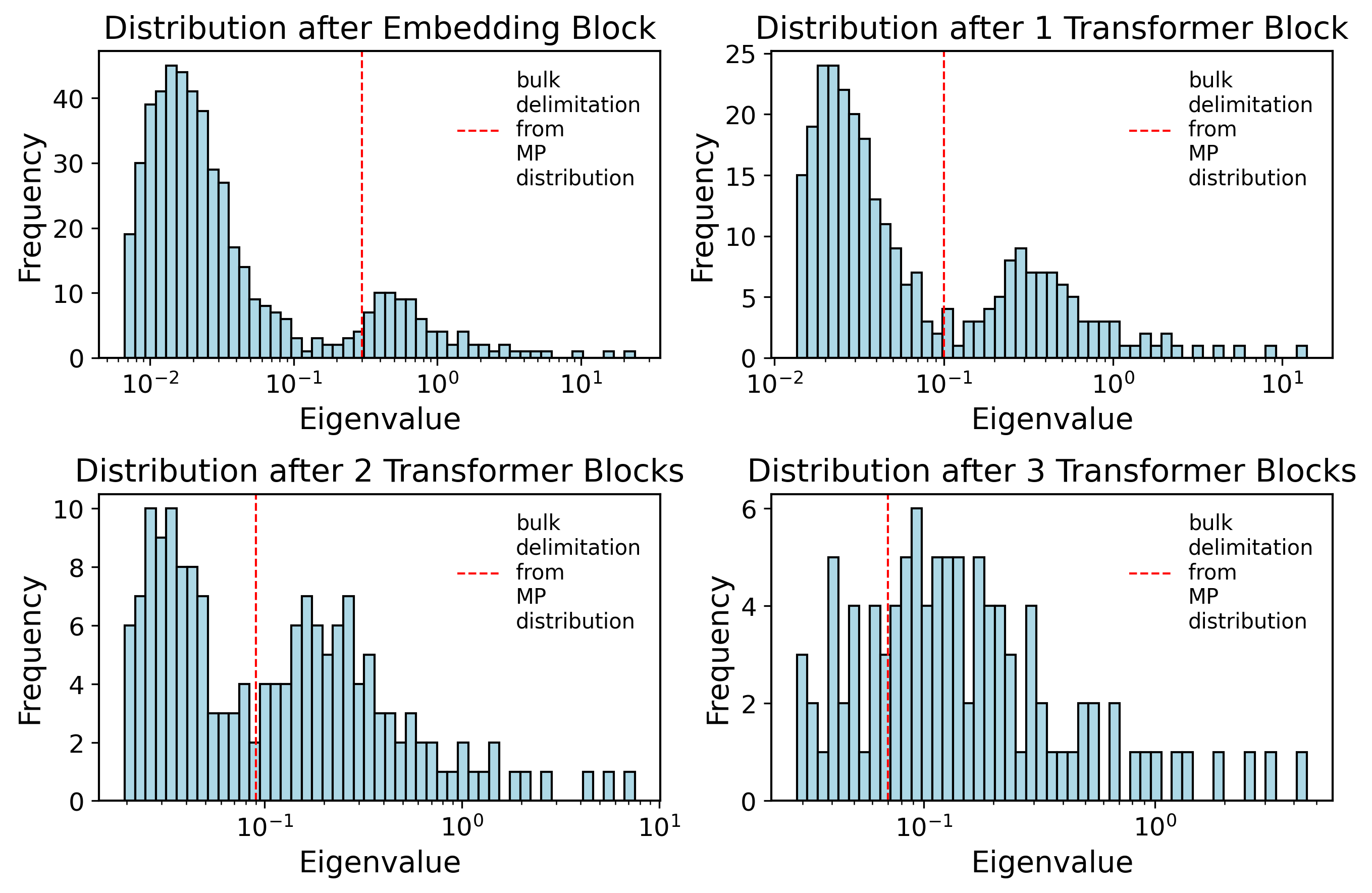}    
    \caption{The empirical eigenvalue distribution of the activation matrices computed on the calibration dataset for BERT-base on the SST dataset.}\vspace{-20pt}
    \label{fig:eig_distribution}
\end{figure}

\vspace{5pt}
\noindent\textbf{\textit{Why RMT-based Model Distillation is Theoretically Justified}}  
Although embeddings from large language and vision–language models are generated by deterministic networks, their covariance structures in high dimensions exhibit statistical regularities that can be described by RMT \cite{zumbach2011empirical}. The eigenvalue spectrum of empirical covariance matrices typically consists of a \textit{bulk}, following the MP distribution, and a few isolated \textit{spikes} corresponding to structured, task-relevant signals. This separation provides a principled criterion for distinguishing meaningful representations from random variations which forms the basis of our method.  

For symmetric random matrices with i.i.d.\ entries of mean zero and variance $\sigma^2$, the Wigner semicircle law \cite{wigner1958distribution} describes the eigenvalue density and defines the noise floor of random fluctuations. For sample covariance matrices $\mathbf{C}=\tfrac{1}{n}\mathbf{X}^\top \mathbf{X}$ with $\mathbf{X}\in\mathbb{R}^{n\times p}$ and variance $\sigma^2$, the eigenvalue distribution converges to the MP law with support $[\lambda_-,\lambda_+] = [\sigma^2(1-\sqrt{c})^2,\ \sigma^2(1+\sqrt{c})^2]$, where $c=p/n$. Eigenvalues inside this bulk reflect random variation, while outliers indicate structure. The spiked covariance model \cite{johnstone2001distribution} formalizes this separation, showing that if signal strength exceeds the Baik–Ben Arous–Péché (BBP) threshold $\lambda_{\text{BBP}}=\sigma^2(1+\sqrt{c})$ \cite{baik2005phase}, some eigenvalues detach from the MP bulk and their eigenvectors align with meaningful, data-dependent directions \cite{benaych2011eigenvectors}. These spikes correspond to causal or semantically structured dimensions that can be isolated.

In modern deep networks, hidden representations are extremely high-dimensional, yet many directions do not contribute to task-relevant information \cite{raghu2017svcca,serre2022pca}. By retaining only eigenvectors linked to outlier eigenvalues, one constructs a projection operator mapping activations to a lower-dimensional, signal-dominant subspace. Unlike PCA, where the cutoff is based on heuristics such as explained variance ratios, RMT provides principled thresholds for separating signal from noise. This RMT-guided filtering discards noisy or redundant dimensions while preserving essential features, yielding narrower yet dense layers. Embedded in an iterative knowledge-distillation loop, it enables compressed models to adapt to the reduced space, maintaining accuracy while reducing memory, latency, and energy consumption.

\begin{figure}[t!]
    \centering
    \includegraphics[width=\linewidth]{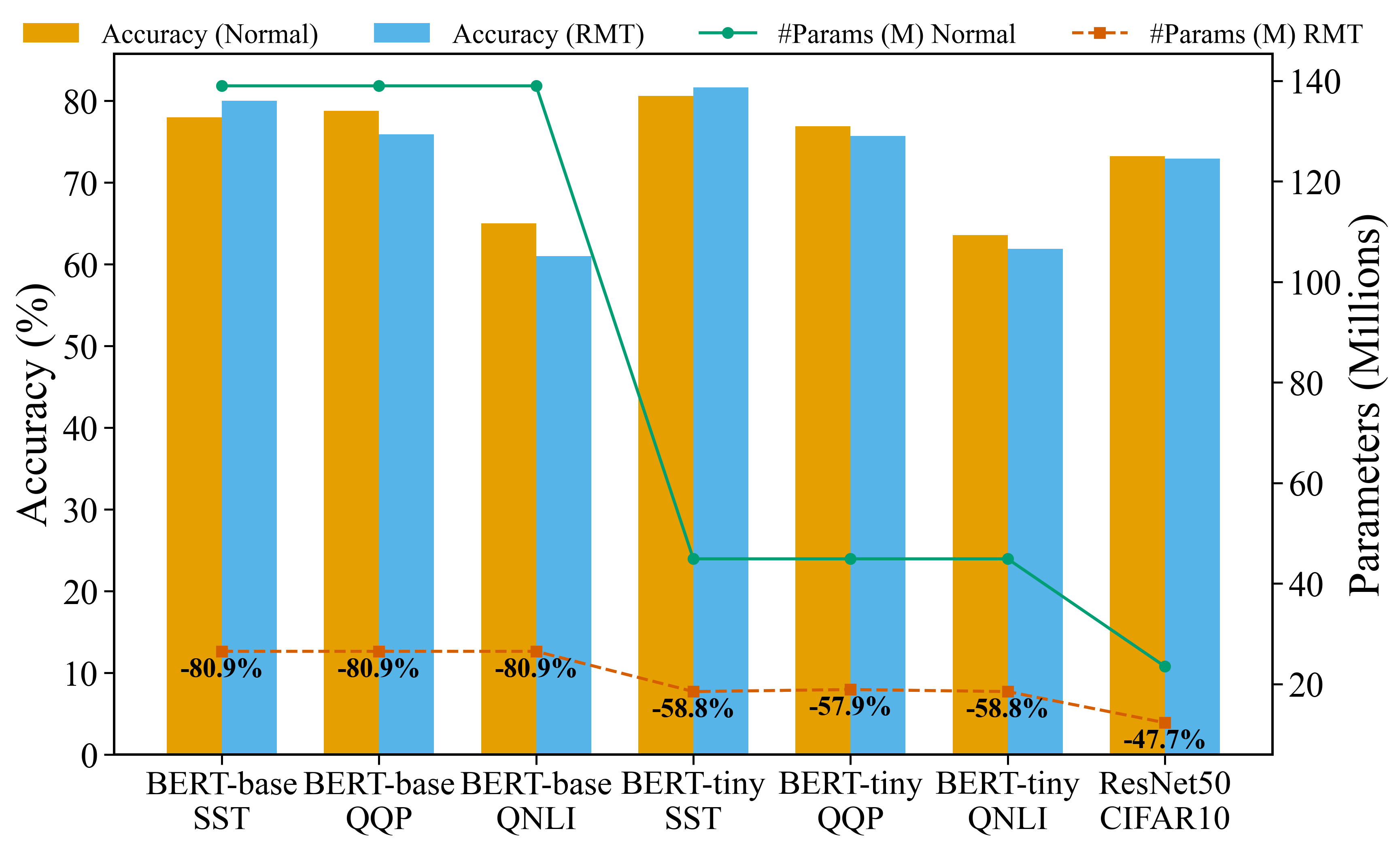}
    \includegraphics[width=\linewidth]{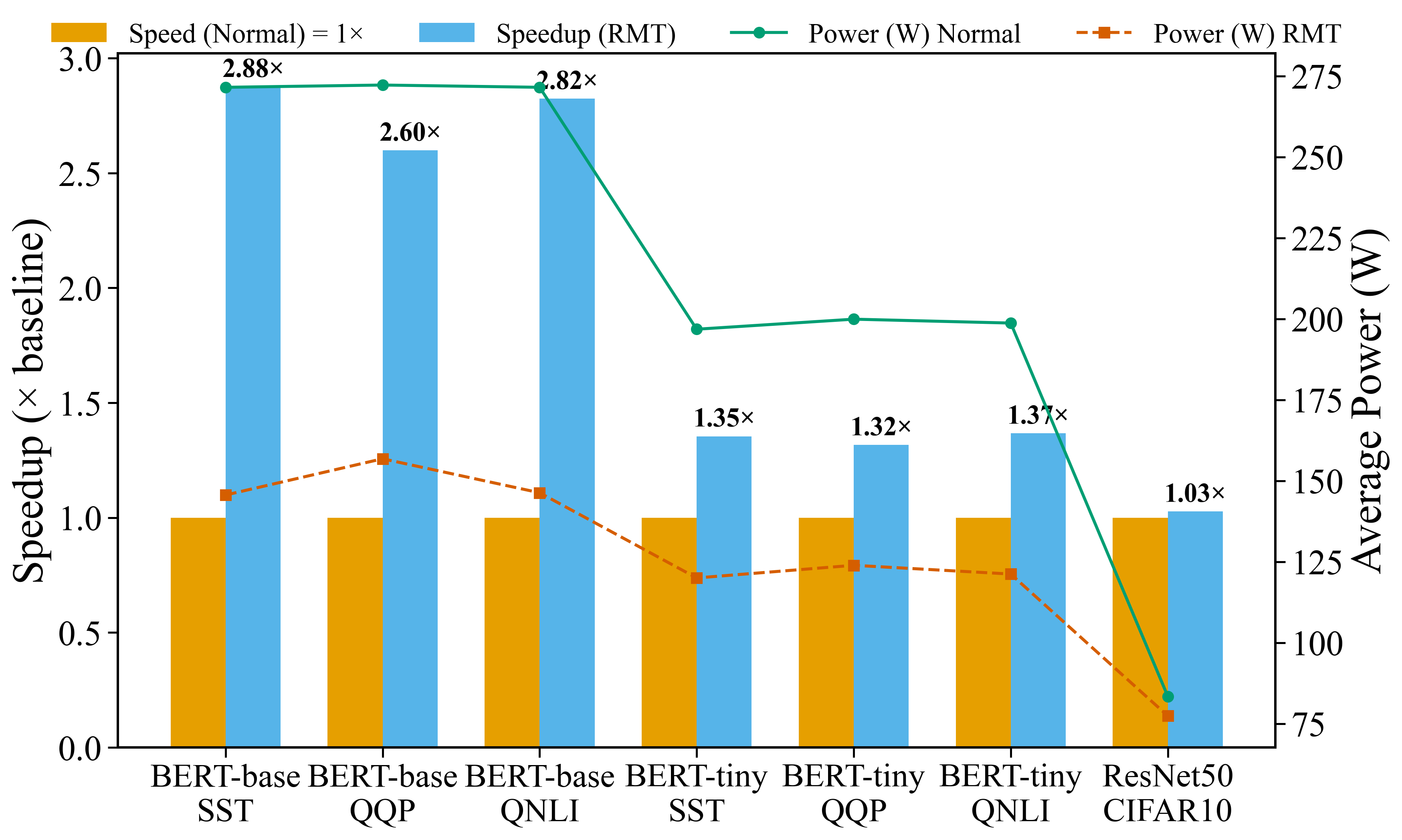}
\caption{\textbf{(a, top)} Accuracy vs. parameter reduction and \textbf{(b, bottom)} power consumption vs. inference speedup on GLUE datasets ($\sigma^2 =$ median eigenvalue, initial quantile = 50\%).}\vspace{-10pt}
    \label{fig:par_acc}
\end{figure}
\section{Results and Discussion}
Fig.~\ref{fig:eig_distribution} shows the evolution of empirical eigenvalue spectra of activation matrices computed on calibration data from different depths of BERT-base trained on SST (GLUE). After the validation accuracy threshold is reached, activations encode structured information rather than random noise, and the spectra deviate markedly from the Wigner semicircle law. In the embedding block, most eigenvalues cluster near zero, with only a few outliers exceeding the Marchenko–Pastur bulk (red dashed line), indicating that most directions are noise-dominated while a small subset carries meaningful signal. Deeper layers exhibit broader spectra with smoother decay and a larger fraction of eigenvalues above the bulk edge, reflecting the accumulation and propagation of task-relevant information. This trend is consistent with the idea that later representations are more specialized and semantically structured. Importantly, the bulk cutoff is computed adaptively from the data using RMT rather than a fixed threshold as in PCA, ensuring that only statistically significant causal directions are retained at each step. This adaptive criterion supports our iterative approach: early layers can be compressed aggressively, while deeper layers with richer spectra need conservative reduction, as their distributions deviate increasingly from the Wigner Semicircle Law. 

Fig.~\ref{fig:par_acc}a compares accuracy and parameter counts before and after RMT-based iterative compression. Across all model–dataset pairs, accuracy remains within 2–3 percentage points of the baseline despite substantial reduction, and in cases such as BERT-base on SST, performance even improves slightly—likely due to removal of noisy or redundant parameters. Parameter counts drop sharply, with BERT-base reduced by about 80\% while retaining near-original accuracy, highlighting its overparameterization and the ability of RMT-based reduction with self-distillation to preserve informative directions. BERT-tiny still shows 58\% reduction, suggesting redundancy even in smaller models, though with less dramatic impact. ResNet-50 exhibits the smallest reduction, around 48\%, reflecting its more compact convolutional structure. Overall, the results indicate that larger, more overparameterized models benefit most from statistically guided compression, while smaller architectures are closer to their efficiency limits.

\begin{figure}[t!]
    \centering
\includegraphics[width=\linewidth]{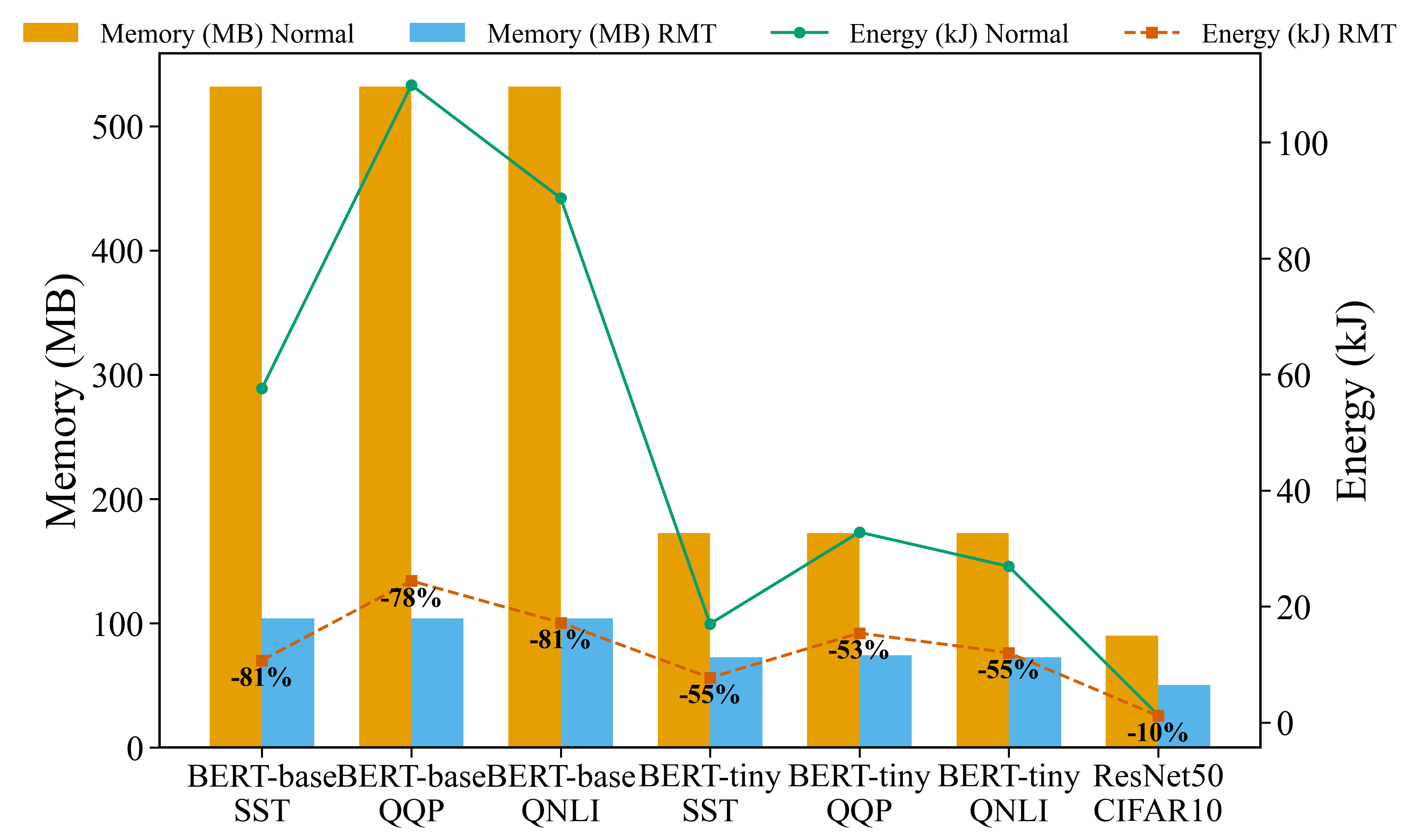}
    \caption{Comparison of memory on disk and energy efficiency for all models and GLUE datasets, $\sigma^2 = Median Eigenvalue$, initial quantile = 50\%}\vspace{-7pt}
    \label{fig:mem_ene}
\end{figure}

Fig.~\ref{fig:par_acc}b shows inference speedup and average power consumption after compression. BERT-base achieves the largest gains, with nearly 3× faster inference on SST and QNLI and slightly lower but still substantial gains on QQP. These improvements stem from the sharp parameter reduction and smaller intermediate representations, which cut both computation and data movement. BERT-tiny shows more modest speedups of 1.3–1.4×, reflecting fixed overheads that dominate smaller models. ResNet-50, already compact, improves only marginally (~1.03×), indicating limited latency benefits from further reduction. In the figure, power consumption consistently decreases across models, most notably for BERT-base where lower compute demand reduces sustained power draw. For BERT-tiny and ResNet-50 the drop is less pronounced, likely because fixed hardware and memory access costs contribute a larger share of energy use.

Fig.~\ref{fig:mem_ene} compares memory footprint and total energy consumption before and after compression. All models show substantial memory savings, with BERT-base dropping from 532~MB to just over 100~MB ($\sim$80\%) and BERT-tiny halving to ~72~MB. These reductions mirror the parameter savings, as fewer weights translate directly to smaller model files. ResNet-50, starting from a smaller size with less redundant structure, shows more modest gains. Energy consumption during inference also decreases consistently, most sharply for BERT-base, which achieves over 5× reduction due to shorter execution and lower sustained power draw. BERT-tiny follows the same trend with smaller absolute savings, while ResNet-50 shows the least change. 

\begin{figure}[t!]
    \centering
\includegraphics[width=\linewidth]{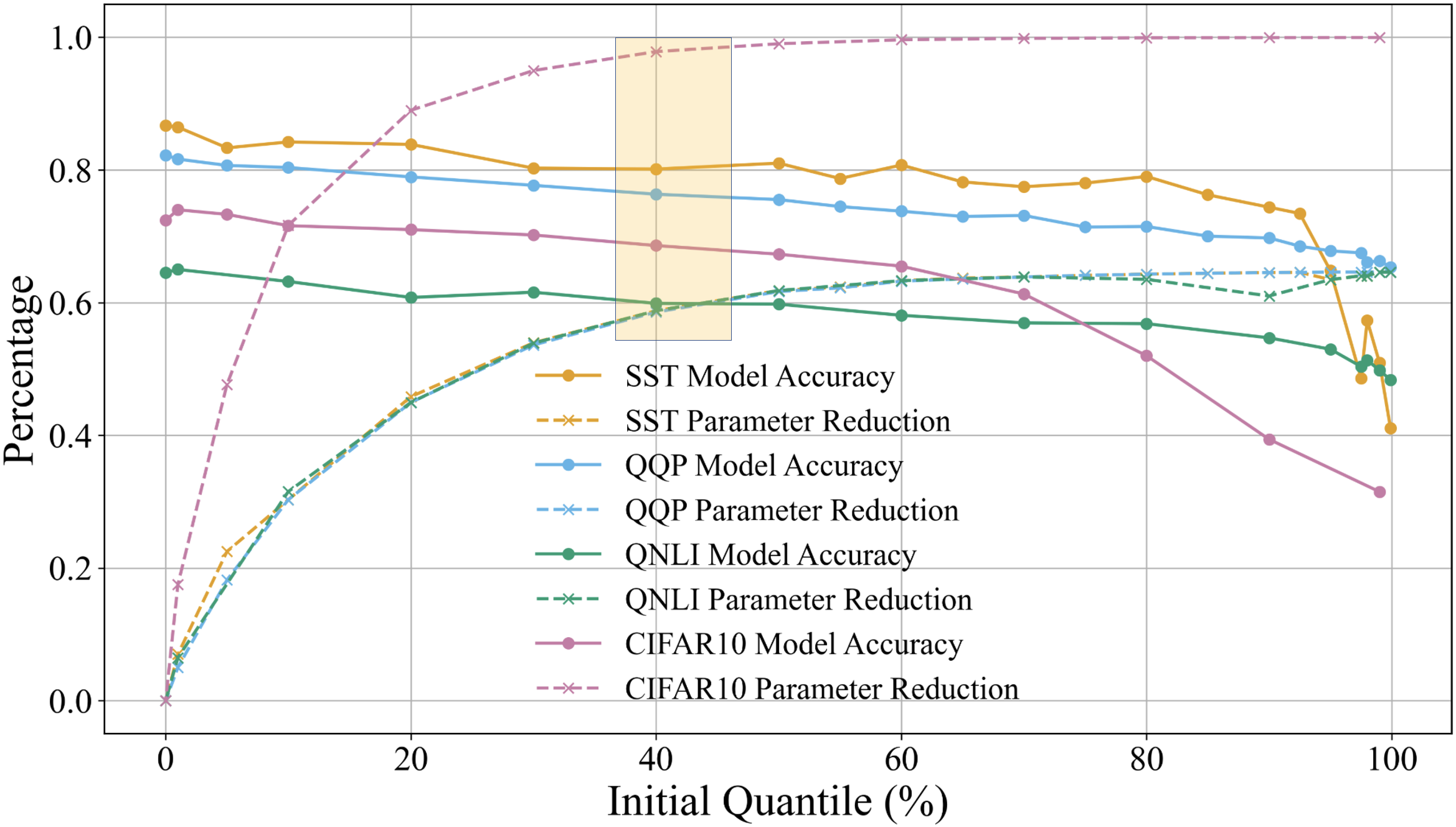}
\caption{Accuracy–reduction tradeoff for BERT-base (GLUE) and ResNet-50 (CIFAR) as a function of the eigenvalue quantile used to initialize $\sigma^2$. The x-axis shows the quantile, the y-axis shows accuracy (decreasing) and parameter reduction (increasing). The best balance occurs near 40\%.}\vspace{-10pt}
    \label{fig:Q_comparison}
\end{figure}

\vspace{5pt}
\noindent\textbf{Ablation Study:} Fig.~\ref{fig:Q_comparison} shows the trade-off between accuracy and parameter reduction when varying the quantile for $\sigma^2$ initialization in BERT (GLUE) and ResNet (CIFAR-10). At low quantiles, accuracy remains close to the baseline but reductions are limited; at high quantiles, aggressive pruning causes sharp accuracy loss. The best balance occurs around the 40\%–50\% quantile, near the median, where both architectures retain high accuracy with substantial compression. This regime is particularly effective for reducing inference time and energy while preserving performance. ResNet offers greater reduction potential, while BERT is constrained by fixed embedding and token projection layers, which dominate beyond the 40\% quantile and limit further compression without accuracy degradation.

\begin{table}[t!]
\centering
\small
\setlength{\tabcolsep}{4.5pt}
\begin{tabular}{llcc}
\toprule
\textbf{Model} & \textbf{Method} & \textbf{Red.} & \textbf{Acc.} \\
\midrule
\multirow{4}{*}{\textbf{BERT-base (GLUE)}} 
& RMT-KD     & \textbf{80.9\%} & \textbf{+1.8\%} \\
& DistilBERT & 42.7\% & +0.2\% \\
& Theseus    & \underline{48.3\%} & \underline{+0.6\%} \\
& PKD        & 40.5\% & -1.0\% \\
\midrule
\multirow{4}{*}{\textbf{BERT-tiny (GLUE)}} 
& RMT-KD     & \textbf{58.8\%} & \textbf{+1.4\%} \\
& DistilBERT & \underline{54.8\%} & \underline{+0.4\%} \\
& Theseus    & 53.0\% & +0.1\% \\
& PKD        & 50.1\% & -0.8\% \\
\midrule
\multirow{4}{*}{\textbf{ResNet-50 (CIFAR-10)}} 
& RMT-KD     & \textbf{47.7\%} & \textbf{+0.7\%} \\
& AT         & 42.2\% & +0.4\% \\
& FitNet     & 40.6\% & +0.2\% \\
& CRD        & \underline{45.4\%} & \underline{+0.6\%} \\
\bottomrule
\end{tabular}
\caption{Comparison of KD methods. Theseus = BERT-of-Theseus, PKD = Patient Knowledge Distillation, AT = Attention Transfer, FitNet = Hints for Thin Deep Nets, CRD = Contrastive Representation Distillation.}
\label{comparison_table}
\end{table}

Table~\ref{comparison_table} compares RMT-KD with state-of-the-art distillation and compression methods across NLP (BERT-base, BERT-tiny on GLUE) and CV (ResNet-50 on CIFAR-10). RMT-KD achieves substantial parameter reductions (up to 80.9\%) while consistently improving accuracy, outperforming specialized baselines such as TinyBERT, Theseus, and CRD. This advantage arises from its ability to dynamically identify and retain only the most causal components of hidden representations. Unlike heuristic pruning or rank selection, RMT-KD leverages Random Matrix Theory: eigenvalues beyond the Marchenko–Pastur threshold mark informative directions, enabling compression driven by rigorous statistical principles rather than ad hoc cutoffs.

\section{Conclusion}
We introduce RMT-KD, an iterative distillation framework that combines layer-wise RMT analysis with self-distillation to compress deep models while preserving accuracy. Unlike pruning or heuristic truncation, RMT-KD offers a statistically grounded, data-driven rule for dimensionality reduction that retains dense causal structure. Experiments on BERT (GLUE) and ResNet (CIFAR-10), it achieves up to 80\% parameter reduction with only 2\% accuracy loss, 2.8× faster inference, and 80\% lower energy use. The models remain hardware-efficient since projections are dense and avoid sparse kernels. Gains are largest for overparameterized transformers, while smaller models like ResNet-50 show modest improvements, reflecting proximity to their efficiency frontier. Performance also depends on calibration subset quality, which may limit robustness under distribution shift.


\vfill
\pagebreak
\newpage

\section*{Acknowledgment}
This work was supported in part by a Gift funding from Intel, by CogniSense, one of the seven SRC/DARPA JUMP2.0 Centers, NSF CAREER Award \# 2046435. 

\bibliographystyle{IEEEbib}
\bibliography{refs}

\end{document}